\documentclass[12pt]{article}
\usepackage{amssymb}
\usepackage{amsmath}
\usepackage{apacite}
\usepackage{latexsym}
\usepackage{epsfig}
\usepackage{graphicx}
\usepackage{subfigure}
\usepackage{wasysym}
\usepackage{threeparttable}
\usepackage{natbib}
\usepackage{color}
\usepackage{epstopdf}
\usepackage{bm}
\usepackage{float}
\usepackage{todonotes}
\usepackage{verbatim}

\usepackage{hyperref}

\newcommand{\E}[1]{{\mbox{E}}\left[#1\right]}

\def\dfrac#1#2{{\displaystyle{#1\over#2}}}

\def\boxit#1{\vbox{\hrule\hbox{\vrule\kern6pt
          \vbox{\kern6pt#1\kern6pt}\kern6pt\vrule}\hrule}}

\def\bse{\begin{eqnarray*}}
\def\ese{\end{eqnarray*}}
\def\be{\begin{eqnarray}}
\def\ee{\end{eqnarray}}
\def\bq{\begin{equation}}
\def\eq{\end{equation}}
\def\bse{\begin{eqnarray*}}
\def\ese{\end{eqnarray*}}

\begin{document}

\thispagestyle{empty} 
\baselineskip=28pt

\textbf{Running Title: Bispectrum-based Nonlinear Time Series Classification}

\begin{center}
{\LARGE{\bf Nonlinear Time Series Classification Using Bispectrum-based Deep Convolutional Neural Networks}}

\end{center}

\baselineskip=12pt

\vskip 2mm
\begin{center}
Paul A. Parker\footnote{(\baselineskip=10pt to whom correspondence should be addressed)
Department of Statistics, University of Missouri,
146 Middlebush Hall, Columbia, MO 65211-6100, paulparker@mail.missouri.edu},
   Scott H. Holan\footnote{\baselineskip=10pt Department of Statistics, University of Missouri,
146 Middlebush Hall, Columbia, MO 65211-6100, holans@missouri.edu}\,\footnote{\baselineskip=10pt U.S. Census Bureau, 4600 Silver Hill Road, Washington, D.C. 20233-9100, scott.holan@census.gov},
 and Nalini Ravishanker\footnote{\baselineskip=10pt
Department of Statistics, University of Connecticut,
215 Glenbrook Road, Storrs, CT 06269-4120, nalini.ravishanker@uconn.edu}
\\
\end{center}
%
%
%
%
\vskip 4mm

\baselineskip=12pt

%

%
%
%

\baselineskip=12pt
\par\vfill\noindent
\par\medskip\noindent
\clearpage\pagebreak\newpage \pagenumbering{arabic}
\baselineskip=24pt

\begin{center}
{\bf Abstract}
\end{center}

Time series classification using novel techniques has experienced a recent resurgence and growing interest from statisticians, subject-domain scientists, and decision makers in business and industry. This is primarily due to the ever increasing amount of big and complex data produced as a result of technological advances. A motivating example is that of Google trends data, which exhibit highly nonlinear behavior. Although a rich literature exists for addressing this problem, existing approaches mostly rely on first and second order properties of the time series, since they typically assume linearity 
of the underlying process.  Often, these are inadequate for effective classification of nonlinear time series data such as Google Trends data. Given these methodological deficiencies and the abundance of nonlinear time series that persist among real-world phenomena, we introduce an approach that merges higher order spectral analysis (HOSA) with deep convolutional neural networks (CNNs) for  classifying time series. The effectiveness of our approach is illustrated using simulated data
and two motivating industry examples that involve Google trends data and electronic device energy consumption data.

	\section{Introduction}

Time series classification is a general task that can be useful across many subject-matter domains and applications. The overall goal is to identify a time series as coming from one of possibly many sources or predefined groups, using labeled training data. That is, in this setting we conduct supervised learning, where the different time series sources are considered known. For example, \citet{hol10} use signals to classify animal behavior, and \citet{lines11} classify household devices based on electricity consumption time series. A closely related problem is time series clustering, where the goal is still to group a collection of time series by their corresponding sources, but no labels for the true sources are known. This constitutes an exercise in unsupervised learning. Time series clustering is often conducted using either hierarchical clustering methods \citep{kaki98, kumar02}, or a $k$-means approach \citep{vlach03, gullo12}. \cite{agha15} give a review of many modern approaches to time series clustering. Although the clustering techniques themselves may not be pertinent in a supervised learning setting, oftentimes data processing or feature extraction steps may be helpful in both the supervised and unsupervised realms. For example, \citet{shumway03} demonstrate that the construction of time-varying spectra can be of use for both classification and clustering of time series.

Although some statistical models such as hidden Markov models may be used to classify time series \citep{chai01}, much of the modern research on time series classification and prediction has come from the machine learning community. For instance, \citet{medeiros19} explore the use of random forests in order to forecast inflation levels. Many of these machine learning methods perform quite well for prediction, but lack some properties that are critical to the field of statistics such as uncertainty quantification, the ability to perform inference, and feature extraction. For example, a distance summary measure such as dynamic time warping can be used in conjunction with nearest neighbor methods to classify time series \citep{jeon11}. Deep learning methods can also be used to classify time series. \citet{fawaz19} review and compare many modern deep learning methods that have been used for time series classification, including convolutional neural networks and echo state networks. \citet{heaton17} demonstrate how deep learning methods may be used for prediction and classification with application to construction of stock portfolios in finance.

Although many of the methods discussed so far consider the series in the time domain, it is also possible to use frequency domain analysis to classify time series. For example, \citet{holan12} use a time-frequency representation to classify periods of economic expansion and recession. \citet{hol18} review many frequency domain methods that may be used to both classify and cluster time series. For nonstationary time series, short-time Fourier transforms may be necessary, such as those used by \citet{hol10} and \citet{shumway03}. Similarly, for nonlinear time series, Higher Order Spectral Analysis (HOSA) may be necessary to capture higher moment properties. \citet{har13} provide an example of HOSA applied to the problem of time series clustering by utilizing various distance metrics in coordination with a hierarchical clustering algorithm, while \citet{har17} discussed clustering nonstationary, nonlinear time series.

To the best of our knowledge, there is no literature available on statistical \textit{classification} of time series using HOSA, particularly for time series that occur in business and industry.
One difficulty may be that the high-dimensionality associated with HOSA techniques can be problematic for many model-based approaches. We have found that simple dimension reduction approaches such as principal components analysis (PCA) 
are generally not sufficient for dealing with this problem. In order to fill this gap in the literature, we develop a HOSA method relying on deep learning that can be used to accurately classify time series. In addition, we use a type of Bayesian deep learning that allows for uncertainty quantification. This provides many of the uncertainty quantification benefits associated with Bayesian modeling, while accommodating the high-dimensional data structure 
corresponding to HOSA, and avoiding the need for costly Markov chain Monte Carlo (MCMC) techniques. Furthermore, our method utilizes a 
variant of feature extraction that can be used to perform inference by identifying the key frequencies used to determine each class probability. As such, this method is useful as a statistical tool for classifying nonlinear time series.

The 
outline of this article 
follows. In Section~\ref{sec:review}, we review the methodological work necessary to construct our proposed method.  This includes a specific type of HOSA based on the bispectrum, CNNs, and the dropout procedure. 
We describe our proposed method in Section 3, where we also discuss a  competing frequency domain method that does not rely on HOSA.  A simulation study is constructed in Section~\ref{sec:sim} in order to evaluate our proposed method. In Section~\ref{sec:app}, we illustrate our method on two example applications. The first involves the classification of Google trends data, and the second involves the classification of household devices based on electricity consumption as discussed in \citet{lines11}. Finally, we give a discussion on the advantages of our proposed methods as well as some insight into possible future work in Section~\ref{sec:disc}.

	\section{Review of Model Components}\label{sec:review}
	
	In order to present our methodology, we first review the necessary components of the model. The first component is the bispectrum, which is used as a form of feature engineering for the raw time series. We then discuss CNNs, which are the main building block of our model. Finally, we consider dropout which may be used as a form of regularization as well as a means for uncertainty quantification.

		\subsection{Bispectrum}

		Traditional  spectral domain analysis resulting from the Fourier transform of the autocorrelation function 
enables classification and clustering of stationary time series only based on their second-order properties. While this can be sufficient for distinguishing between linear time series, the use of higher-order spectral analysis (HOSA) is 
attractive for classifying or clustering nonlinear time series. Higher-order spectra (HOS) are Fourier transforms of  
third-moments or higher-order moments. Properties of HOS have been discussed in  \cite{randvn1965}, \cite{vanness1966},
\cite{bandr1967}, or \cite{hinich1982}, among others.

Let $\{\varepsilon_t: t \in {\cal{Z}}\}$ represent a white noise
process with variance $\sigma^2_\varepsilon$.  
%
%
For a zero-mean, third-order stationary series $\{x_t\}$, the autocovariance
and third-order moment functions are defined as
$$
\gamma_v = \E{X_t X_{t+v}}  {\mbox{and}} \quad
\gamma_{u,v} = \E{X_t X_{t+u}X_{t+v}},
$$
respectively. 
%

If     
$\sum_{u=-\infty}^\infty\sum_{v=-\infty}^\infty\!|\gamma_{u,v}| <
\infty$, the bispectral density function is
$$
I(\omega_1, \omega_2) =
\sum_{u=-\infty}^\infty\sum_{v=-\infty}^\infty\!\gamma_{u,v}
  e^{-2\pi i(u\omega_1+v\omega_2)}, \quad {\mbox{for $(\omega_1,\omega_2) \in
 [0,1] \times [0,1],$}}.
$$

%
%

Taking note of the symmetries in $I(\omega_1,\omega_2)$ across the
unit square, the principal domain of $I(\omega_1,\omega_2)$ and
$Z(\omega_1,\omega_2)$ can be shown to be the triangle with boundaries
${\cal{D}} = \{(\omega_1,\omega_2): 0 \le \omega_2 \le \omega_1 \le
0.5,~ 2\omega_1 \le (1-\omega_2)\}$.

Let $x_t,~t = 1, 2, \ldots, n$ represent a mean-corrected realization
of the process $\{X_t\}$, and without loss of generality, assume
$\E{X_t} = 0$.  Then the sample autocovariance and third-moment
functions are
$$
\hat\gamma_v = \frac{1}{n}\sum_{t=1}^{n-v}x_t x_{t+v}
\quad {\mbox{and}} \quad
\hat\gamma_{u,v} = \frac{1}{n}\sum_{t=1}^{n-s}x_t x_{t+u}x_{t+v},
$$
where $s = \max\{0,u,v\}$. Let $\lambda(\tau)$ represent a symmetric
lag window with $\lambda(0) = 1$, and $\lambda(\tau_1,\tau_2)$
represent a two-dimensional lag window satisfying
$$
\lambda(\tau_1,\tau_2) = \lambda(\tau_2,\tau_1) =
\lambda(-\tau_1,\tau_2-\tau_1) = \lambda(\tau_1-\tau_2,-\tau_2) =
\lambda(\tau_1)\lambda(\tau_2)\lambda(\tau_1+\tau_2),
$$
corresponding to the symmetries in $I(\omega_1,\omega_2)$ over the
unit square.  Define the natural frequencies $\omega_j =
(j-1)/n,~j=1,2,\ldots,[n/2]+1$, where $[\cdot]$ represents greatest
integer value, and consider $(\omega_j, \omega_k) \in D = \{(j,k):
1 \le k \le j \le [n/2]+1,~2j+k-3 \le n\}$, corresponding to ${\cal{D}}$.
Then the sample spectral and bispectral density functions are defined as
\begin{eqnarray}
\hat I(\omega_j) & = & \sum_{v=-M}^M\!\lambda\left(\frac{v}{M}\right)
       \hat\gamma_v e^{-2\pi i v\omega_j}   \label{spec} \\
\hat I(\omega_j,\omega_k) & = & \sum_{u=-M}^M\sum_{v=-M}^M\!
       \lambda\left(\frac{u}{M}, \frac{v}{M}\right)
       \hat\gamma_{u,v} e^{-2\pi i (u\omega_j+v\omega_k)} \label{bispec}
\end{eqnarray}

\cite{har13} use a normalized sample bispectral density to cluster nonlinear time series based on various distance metrics.  Their work utilizes the smoothed sample bispectral density, which is defined by (\ref{bispec}).
%
%
Because this sample bispectral density reflects third-moment properties, it may be used to classify nonlinear time series.  In contrast to their approach, 
we consider the unsmoothed bispectrum, noting that our approach provides an implicit smoothing.  Notably, the dimensionality of the sample bispectrum can be quite high, and thus dimension reduction may be required.
		
		\subsection{Convolutional Neural Networks}
\begin{figure}
			\begin{center}
                \includegraphics[width=130mm, scale=0.7]{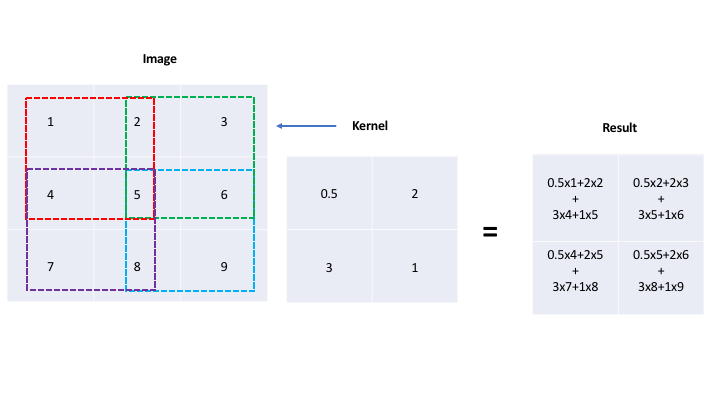}
               \end{center}
                 \caption{\baselineskip=10pt An illustrated example of the 2-d convolution operation. The kernel is applied to each $2 \times 2$ group of pixels (denoted by the different colored dashed lines) within the original image. The image values are multiplied by the corresponding kernel weights, which results in a new $2 \times 2$ image.}
                 \label{fig:conv}
            \end{figure}

		Convolutional neural networks (CNNs), introduced by \citet{lecun89}, are a powerful tool for processing image data. The basis of these networks is the 2-dimensional discrete convolution operator, defined as
		
		\begin{equation}
		    f(x,y) * g(x,y) = \sum_{i=- \infty}^{\infty} \sum_{j=- \infty}^{\infty} f(i,j) g(x-i,y-j),
		\end{equation} where $f$ is a kernel function and $g$ is the image for which convolution is being performed. The values of $x$ and $y$ denote the indices of pixels in the image. An illustration of the discrete convolution operation can be found in Figure \ref{fig:conv}.
		
		Convolution can be thought of as creating a new image by replacing each pixel with a weighted average of nearby pixels. The weights are determined by the kernel, which is often referred to as a filter in the deep learning literature (for example, see \citet{zeiler14}). By including convolution as one component of a deep model, it is possible to use gradient descent or some other optimization technique to estimate the values of the weights. Note that, in practice, with deep learning models, it is typical to use stochastic gradient descent, whereby each iteration of the gradient descent algorithm is based on only a sample of the training data. It is also typical to use many filters in a single convolution layer. Let $N_{l}^f$ be the number of of filters in convolution layer $l$. Then this layer will consist of $N_{l}^f$ kernels that need to be learned, and application of this convolution layer to a single image will result in $N_{l}^f$ new images. For all of our analysis, we use $N_{l}^f=32$.
		
		Another operation that is commonly used in CNNs is called max-pooling, which works by dividing the image into rectangular subsections, and taking the maximum value within each subsection, which results in a new lower dimensional image. Max-pooling can be thought of as applying the maximum operator over a non-overlapping contiguous grid placed upon the image. Along with dimension reduction, max-pooling is beneficial, because it helps to provide translation invariance of the original image space. \citet{good16} review both the convolution operator and max-pooling, as well as provide details on the optimization techniques commonly used in deep learning.
		
		\subsection{Dropout}
		
		A common tool to add parameter regularization and prevent overfitting in deep models is dropout. Dropout is performed by randomly setting a fixed proportion (known as the dropout rate) of the outputs from a given layer to zero. The dropout rate can be tuned via cross-validation to find the optimal value. Traditionally, dropout has been used to train the model, but at prediction time, dropout is not used, and the parameters are scaled to account for this.
		
		\citet{gal16} take the approach of using dropout both for model training as well as for prediction. In this manner, rather than a single prediction, a distribution of predictions is obtained, since the outputs of each layer being randomly dropped may vary. The authors show that dropout in this way can be interpreted as a variational Bayes approximation to a deep Gaussian process model. This gives a theoretically justified measure of model uncertainty, with little additional computational burden.

			\section{Proposed Methodology}\label{sec:meth}
			
			Using the components outlined in Section~\ref{sec:review}, we present our methodology. In addition to this, we compare to a similar classification technique that only considers the time-series spectrum (i.e. no third-order properties).
			
		\subsection{Bispectrum-based Convolutional Neural Networks}
		
		We introduce the Bispectrum-based Convolutional Neural Network (BCNN) with layers $l=1,\ldots,L$ for classification of time series. Because the bispectrum involves third moment properties, this method outperforms spectral-based classification methods on nonlinear time series.
		
		The input into the BCNN is an array of size $n \times T \times T$, where $n$ is the number of time series in the sample, and $T$ is the length of the time series. This array should consist of the raw sample bispectra for the time series in the sample. We do not use the smoothed bispectra as our input, as smoothing is a convolution operation, which can be learned by the neural network. Note that if the time series have different lengths, then they may be zero padded to the same length. Input images are always scaled such that each pixel has mean zero and standard deviation of one.
		
		The first two hidden layers in the BCNN are a convolution layer with max pooling, followed by a dropout layer. A nonlinear transformation is applied after each max pooling stage. In our case, we used the rectified linear unit (ReLU) function, defined as $f(x)=\max(0,x)$, which is applied element-wise. Subsequent convolution and dropout layers may be added as well. The number of desired convolution/dropout layers may be data dependent, but we found that two of each layer worked well for our analyses. For the first convolution layer, we use $3 \times 3$ max pooling, and we use $4 \times 4$ max pooling for the second. We also use $3 \times 3$ and $5 \times 5$ kernel sizes for the first and second convolution layers respectively.
		
		The convolution stage of the BCNN is followed by a densely connected hidden layer. The output of time series $i$ through a dense layer indexed by $l$, with $N_l$ hidden units, is defined as
		\begin{equation}
		    \mathbf{v}_i^l = g(\mathbf{W}^l \mathbf{v}_i^{l-1} + \mathbf{b}^l)  \label{outdense}
\end{equation}		
 where $\mathbf{v}_i^{l-1}$ is the output vector from layer $l-1$ with  
dimension $N_{l-1}$ $\mathbf{W}^l$ is an $N_l \times N_{l-1}$ parameter matrix, and $\mathbf{b}^l$ is a $N_l$-dimensional vector of intercept (also called bias) terms. The function $g(\cdot)$ is some nonlinear activation function, where again, we use the ReLU function. Note that the dense layer simply takes an input, $\mathbf{v}^{l-1}$, and performs a linear transformation followed by a nonlinear transformation. As with the convolution layers, the number of densely connected hidden layers and their corresponding number of hidden units may be varied. We found that for our work, one dense layer of size 8 hidden units worked well. We also follow the last hidden layer with a dropout layer. Although additional dropout layers may be added with varying dropout rates, we found that additional dropout layers did not add much benefit, and thus use only one in order to limit the necessary cross-validation.
		
		The final layer of the BCNN is the output layer. For classification of time series  $i=1,\ldots,n$ into $K$ categories, this layer is defined as
		\begin{align*}
		    \mathbf{v}_i^L &= \hbox{softmax}(\mathbf{z}_i^L) = \frac{\exp(\mathbf{z}_i^L)}{\sum \exp(\mathbf{z_i}^L)} \\
		    \mathbf{z}_i^L &= \mathbf{W}^L \mathbf{v}_i^{L-1} + \mathbf{b}^L.
		\end{align*} The output, $\mathbf{v}_i^L$ is a vector of length $K$ giving the class probabilities for time series $i$. Similar to the hidden dense layers, $\mathbf{W}^L$ is a $K \times N_{L-1}$ parameter matrix, $\mathbf{b}^L$ is a $K$-dimensional bias vector, and $\mathbf{v}_i^{L-1}$ is the output from the previous layer. Note that the $\exp(\cdot)$ function is applied element-wise to the vector $\mathbf{z}_i^L$.
		
		Our BCNN model was fit using Keras \citep{keras}. We used a cross-entropy loss function with the Adam optimizer \citep{kingma14}. Adam is a variant of stochastic gradient descent that uses second moment approximations for more efficient updates. All optimization was done for 20 epochs, using a batch size of 8. These values could be tuned, but we found that our results were not very sensitive to this. 
		
		There are many architectural decisions that may be varied in the BCNN: the number of convolution layers and the number of filters per layer, the kernel sizes, as well as the number of dense layers and the number of hidden units per layer. We found that as a general rule, starting with a small network (a single convolution layer and a single hidden dense layer with relatively few filters/hidden units) and gradually increasing until performance seems to flatten seems to yield good results. Our analysis is meant to be illustrative, so we use the architecture described in this section, but these settings could be further tuned if desired.

		\subsection{Comparison to Spectral-SSVS}
		
		In order to compare our proposed BCNN to methodology that only considers spectral properties, we introduce spectral-stochastic search variable selection (spectral-SSVS). This method is based on dimension reduction and variable selection with a binary outcome, and is a special case of \citet{hol10}. We begin by calculating the periodogram, or the unsmoothed version of (\ref{spec}), for each time series. Since the periodogram can be high dimensional, we use a principal components analysis to reduce the feature space. The number of components (PCs) retained should depend on the data, but for our simulation study, we retained 20 PCs.
		
		The spectral-SSVS method then uses a Bernoulli response model (for classification into two classes), with the retained PCs as covariates.
		Dimension reduction results in a new set of features, but each feature in this reduced space is not necessarily correlated to the response of interest.  In this case, the response is the class that each time series belongs to. We require some way to select the important features in terms of predicting the class associated with a given time series, and we recommend Bayesian SSVS as shown by \cite{geo97}. Because our response is binary, we can use data augmentation to form a probit model as done by \cite{alb93}. The full model for $Y_i$, $i=1,\ldots , n$ is
		$$\begin{aligned}
			Y_i &= \begin{cases}
				1; & \text{if $Z_i>0$},\\
    				0;  & \text{if $Z_i \leq 0$},
  			\end{cases} \\
			Z_i | \bm{\beta} & \sim N(\bm{x}_i ' \bm{\beta}, 1) \\
			\beta_j | \gamma_j & \sim \gamma_j N(0, c_j \tau_j^2) + (1-\gamma_j) N(0, \tau_j^2), \quad j=1, \ldots , p \\
			\gamma_j & \stackrel{iid}{\sim} Bern(\pi_j)
		\end{aligned}
		$$
		Here $Y_i = 0$ if the $i$th time series is of the first class and $Y_i = 1$ if the time series is of the second class. The continuous latent variable $Z_i$ is used to relate the response to the linear predictor. The covariates for the $i$th time series, $\bm{x}_i $, as well as $\bm{\beta}$ are $p \times 1$ vectors, in our case, the PCs. All analysis done here will include an intercept term in the construction of the regression coefficients. The prior on $\bm{\beta}$ is the SSVS  prior, where the mixture of normals representation allows for the use of a Gibbs sampler.  The constants $c_j$, $\tau_j$, and $\pi_j$ are all tuning parameters that must be selected by the user. 
		
		The parameter $\gamma_j$ can be seen as an indicator of whether or not the $j$th covariate is ``selected'' as important.  For this reason, we would like $\tau_j$ to be small and $c_j$ to be large so that if $\gamma_j = 0$ then the prior variance on $\beta_j$ will be small and thus $\beta_j$ will be shrunk towards zero.  Cross-validation can be used to tune these parameters. In our case, we standardized all covariates.  We found that for the probit SSVS model given here with standardized covariates, $\tau_j=0.1$, $c_j=10$, $\pi_j = 0.5$ for all $j$ seemed to work well.  Further fine tuning could be done, but we found that this yields minimal gains in prediction accuracy.
		
		Because this method is Bayesian, we gain a full distribution for the selection of each covariate. Specifically, at each iteration of the MCMC, the current value of $\gamma_j$ indicates whether the $j$th covariate is selected.  One could average over all MCMC samples and then refit a model only using the covariates that are selected with probability greater than some threshold.  We can use this model to predict the class of out of sample data.  Another option is to use Bayesian model averaging (BMA) \cite{hoe99}.  At each iteration of the MCMC we can make a prediction for the class of out of sample data using the current values of $\bm{\beta}$. We can then average over the class prediction done at each iteration of the MCMC to get a class probability that takes into account the uncertainty of all parameters.  We use BMA for all class predictions in our simulation study.
		
	    Note that we did attempt to apply SSVS to the PCs generated by the sample bispectra rather than the spectra. However, this methodology did not improve upon spectral-SSVS, likely because nonlinear modeling is required for bispectral features. We do not present those results in this work.

	\section{Simulation Study}\label{sec:sim}

	\cite{har13} use the bispectral periodogram to perform clustering on time series using various distance metrics.  Although their goal was similar to our goal of classification, there are some key differences.  They did not use a model to perform clustering, and thus no training data was used.  They also did not have class labels, resulting in zero loss as long as two time series are put into separate clusters given that they are truly from separate classes.  In our case, if two time series from different classes were both classified into the wrong class, each case would result in a reduction of classification accuracy, even though the two time series were not put into the same class.  We perform a similar simulation study as \cite{har13}, although the two studies cannot be compared on equal footing due to their difference in goals.  Instead of using the median rand index as they did, we use prediction accuracy to represent a similar singular measure of success for the goal of classification. Prediction accuracy is the ratio of the number of correct classifications to the number of classifications performed. We also consider area under the ROC curve (AUC).
	
	Our simulation study considers 7 different nonlinear processes.  For each possible pair of processes, we simulate 200 time series of each process (400 total time series) with 100 time points each. We randomly split the simulated data into a training and validation set, of 200 time series each.  We calculate the bispectral periodogram as inputs into our BCNN model.  After fitting the BCNN model, we generate 100 ensemble predictions on the validation data.  We use the mean predictions over the ensembles to calculate both prediction accuracy and AUC.  The prediction accuracy is obtained from these predictions by noting the number of correct classifications using a probability threshold of $0.5$.  For BCNN, we use two convolution layers with max pooling and ReLU activation, and one hidden dense layer, with 8 hidden units before the output layer, again with ReLU activation. We also use dropout with a rate of $0.1$. We compare out results to the spectral-SSVS model, which is trained and validated on the same data. For spectral-SSVS, we keep 20 PCs, and set $\tau=0.1$, $c=10$, and $\pi = 0.5$. 
	
	The processes considered are the same nonlinear processes considered by \cite{har13}, and include:

	\noindent Three bilinear (BILIN) processes:
	\begin{align*}
	I: x_t &= 0.4 x_{t-1} + 0.4 x_{t-1} \epsilon_{t-1} + \epsilon_t, \\
	II: x_t &= 0.4 x_{t-1} + 0.6 x_{t-1} \epsilon_{t-1} + \epsilon_t, \\
	III: x_t &= -0.2 x_{t-1} + 0.4 x_{t-2} + 0.6 x_{t-1} \epsilon_{t-1} + 0.7 x_{t-2} \epsilon_{t-1} + \epsilon_t, 
	\end{align*}
	\\
	Two self-exciting threshold autoregressive (SETAR) processes:
	\begin{align*}
	IV: \hbox{With } \sigma_{\epsilon}^2 &= 0.003^2, \\
	x_t &= \begin{cases}
		1.2270 + 1.0516 x_{t-1} - 0.8901 x_{t-2} - 0.2149 x_{t-3} + \epsilon_t^{(1)} & \text{if $x_{t-1} \leq 0.5$}, \\
		1.6734 - 0.8295 x_{t-1} + 0.1309 x_{t-2} - 0.0276 x_{t-3} + \epsilon_t^{(2)} & \text{if $x_{t-1} > 0.5$},  
		\end{cases} \\
	V: \hbox{With} \sigma_{\epsilon}^2 &= 0.003^2, \\
	x_t &= \begin{cases}
		0.15 + 0.85 x_{t-1} + 0.22 x_{t-2} - 0.70 x_{t-3} + \epsilon_t^{(1)} & \text{if $x_{t-1} \leq 3.05$}, \\
		0.30 - 0.80 x_{t-1} + 0.2 x_{t-2} + 0.70 x_{t-3} + \epsilon_t^{(2)} & \text{if $x_{t-1} > 3.05$}, 
	\end{cases}
	\end{align*}
	\\
	and two other nonlinear processes:
	\begin{align*}
	VI: & \hbox{ an exponential autoregressive  (EXPAR) model } \\ & x_t = 0.5 x_{t-1} + 1.5 x_{t-1} \exp\{-0.5 x^2_{t-1}\} + \epsilon_t \\
	& \hbox{and} \\
	VII: & \hbox{ a  polynomial  autoregressive  (POLYAR)  model }  \\ 
	& x_t = 0.3452 x_{t-1} + 0.1204 x^2_{t-1} - 0.0994 x_{t-2} + 0.1162 x_{t-1} x_{t-2} + \epsilon_t.
	\end{align*} The (BILIN) models used are those from \cite{rao12},  (SETAR) models from \cite{ton80}, and (EXPAR) model from \cite{jon78}.
	
	The results of this simulation can be found in Table \ref{table:sim}. In most cases, the two methods both work extremely well, indicating that for these processes, the second order properties alone are enough to differentiate them. In the cases where spectral-SSVS outperforms BCNN, the difference in performance is quite small. However, there are three cases  (IV vs. VI, VI vs. IX, and VI vs. X), where BCNN seems to outperform spectral-SSVS by a considerable margin.
	
		\begin{table}[H]
	\begin{center}
		 \begin{tabular}{||c c c c c c||} 
 		\hline
		 Process 1 & Process 2 & BCNN Accuracy & SSVS Accuracy & BCNN AUC & SSVS AUC \\ [0.5ex] 
		 \hline\hline
		 I & II & \textbf{0.80} & 0.76 & \textbf{0.88} & 0.85\\ 
		 \hline
		 I & III & \textbf{0.98} &  0.90 & \textbf{1.00} & \textbf{1.00}\\
		 \hline
		 I & IV & \textbf{1.00}  & \textbf{1.00}  & \textbf{1.00} &  \textbf{1.00}\\
		 \hline
		 I & V & \textbf{1.00}  & \textbf{1.00}  & \textbf{1.00} & \textbf{1.00}\\
		 \hline
		 I & VI & \textbf{0.98}  & 0.97  & \textbf{0.99} & 0.99 \\ 
		 \hline
		 I & VII & 0.84 & \textbf{0.87}  & 0.92 &  \textbf{0.93}\\
		 \hline
		 II & III & \textbf{0.94}  &  0.92 & 0.96 &  \textbf{0.99}\\ 
		 \hline
		 II & IV & \textbf{1.00}  &  \textbf{1.00} & \textbf{1.00} &  \textbf{1.00}\\ 
		 \hline
		 II & V & \textbf{1.00} &  \textbf{1.00} & \textbf{1.00} &  \textbf{1.00}\\
		 \hline
		 II & VI & \textbf{0.96} &  0.94 & \textbf{1.00} &  0.99\\
		 \hline
		 II & VII & \textbf{0.98} &  0.96 & \textbf{1.00} &  0.99\\
		 \hline
		 III & IV & \textbf{1.00}  & 0.98  & \textbf{1.00} & \textbf{1.00}\\ 
		 \hline
		 III & V & \textbf{1.00} & 0.94  & \textbf{1.00} &   \textbf{1.00}\\
		 \hline
		 III & VI & \textbf{1.00} & 0.88  & \textbf{1.00} & 0.92 \\
		 \hline
		 III & VII & \textbf{0.98} & 0.90  & \textbf{1.00} &   0.98\\ 
		 \hline
		 IV & V & \textbf{1.00} &  \textbf{1.00} & \textbf{1.00} &  \textbf{1.00}\\
		 \hline
		 IV & VI & \textbf{1.00} & \textbf{1.00} & \textbf{1.00}  &  \textbf{1.00}\\
		 \hline
		 IV & VII & \textbf{1.00}  & \textbf{1.00} & \textbf{1.00}  &  \textbf{1.00}\\
		 \hline
		 V & VI & \textbf{1.00} & \textbf{1.00}  & \textbf{1.00} &  \textbf{1.00}\\ 
		 \hline
		 V & VII & \textbf{1.00}  & \textbf{1.00} &  \textbf{1.00} & \textbf{1.00} \\
		 \hline
		 VI & VII & \textbf{0.97}  & 0.96  & \textbf{1.00} &  1.00\\
		 \hline
		\end{tabular}
	\end{center}
	\caption{Prediction accuracy and area under the ROC curve for BCNN vs. spectral-SSVS when generating from two nonlinear processes.}
	\label{table:sim}
	\end{table}

	\section{BCNN for Two Business Applications}\label{sec:app}
	
	Using two data sources relevant to industry interests, we illustrate the utility of the BCNN. Comparison is also made to spectrum-based classification techniques. We first consider the classification of Google Trends data, which exhibit highly nonlinear behavior. We also show results for classification of various electronic devices based on electricity consumption.
	
	\subsection{Google Trends Data}
\begin{figure}
				\begin{center}
                \includegraphics[width=180mm]{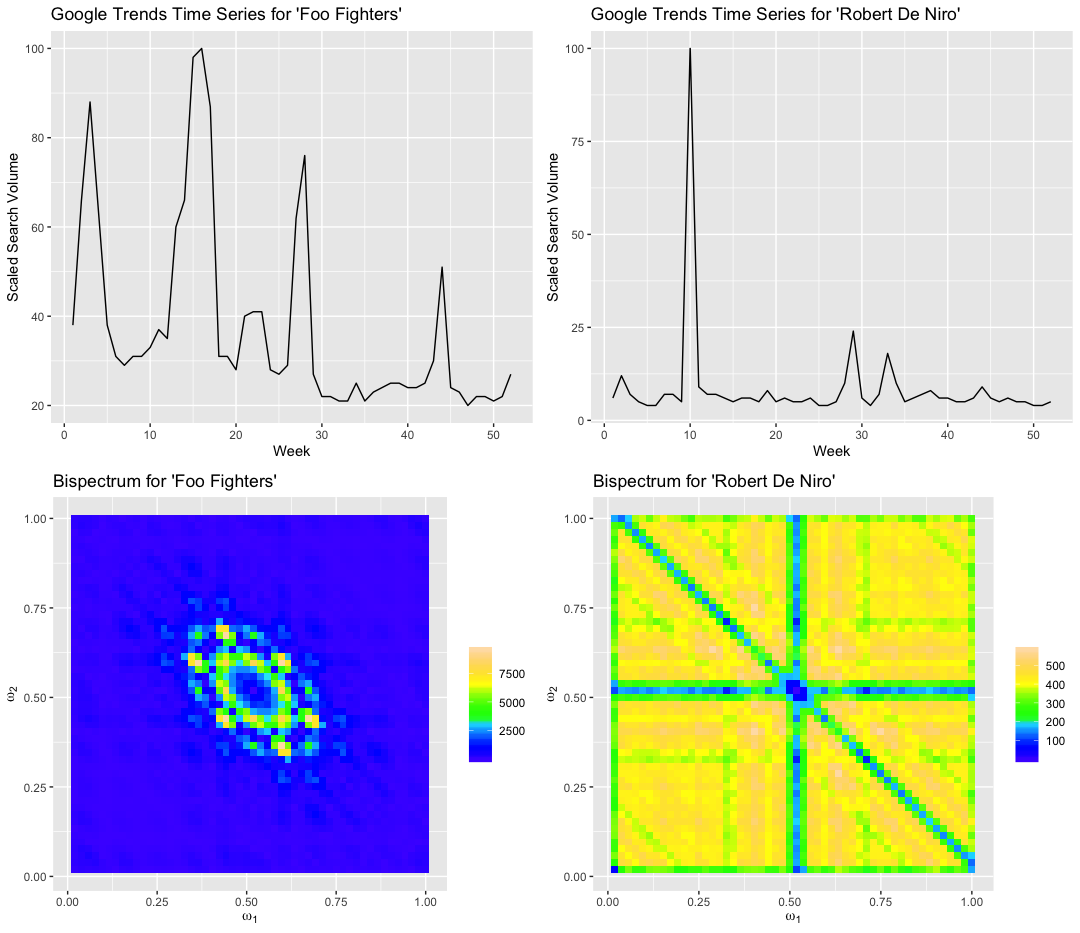}
                \end{center}
                 \caption{\baselineskip=10pt Example bispectral periodograms for two Google Trends time series.}
                 \label{fig:goog}
            \end{figure}

	We apply our methodology to a real nonlinear time series classification problem. Specifically, we obtain weekly Google Trends results for 200 search terms over the 52 week period ending on March 31, 2019. Each time series comes scaled between 0-100, with 100 representing the maximum search volume for that time series during the 52 week period. The search terms consist of the names of 100 different actors, and 100 different music artists, with the goal of classifying the Google Trends time series as either actor or music artist. We display the bispectral periodograms for two example searches in Figure \ref{fig:goog}.

                            				\begin{figure}
              \begin{center}
                \includegraphics[width=150mm]{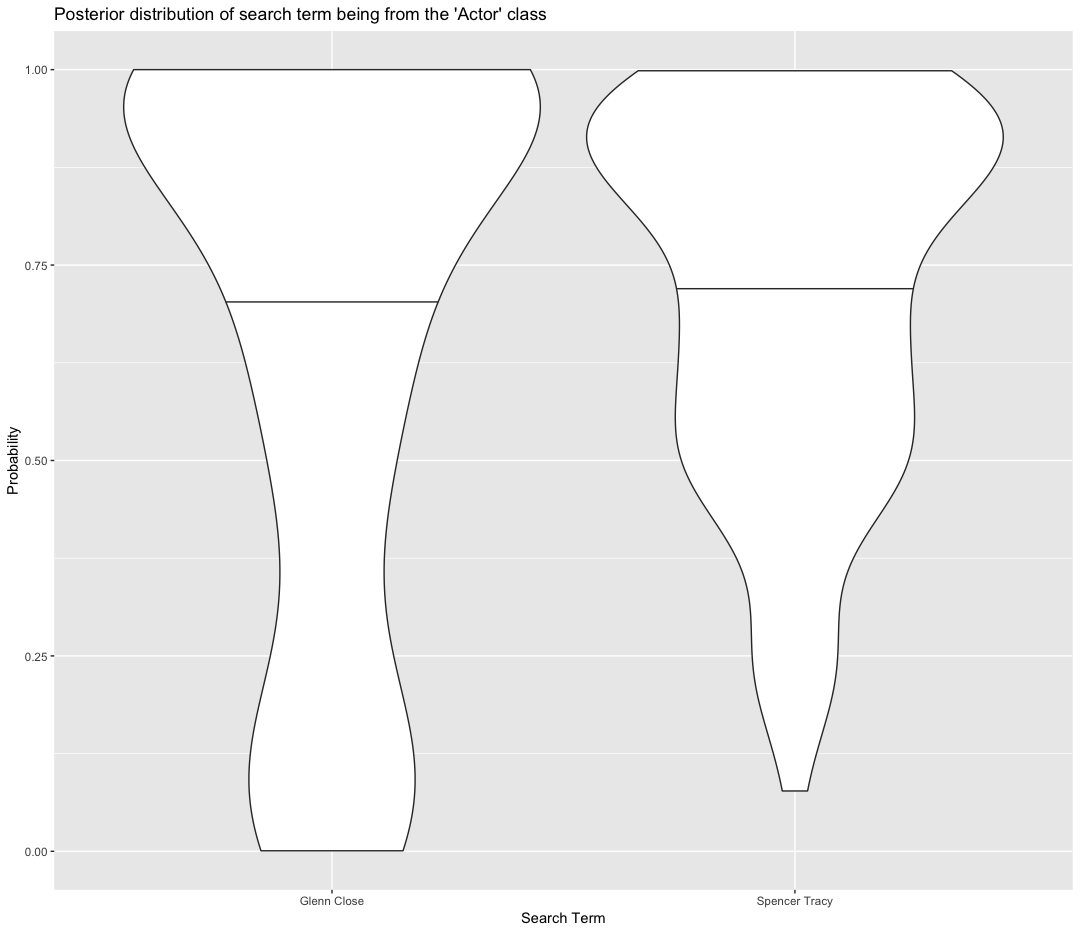}
                 \end{center}
                 \caption{\baselineskip=10pt Posterior probability, with medians, of search time series coming from the `Actor' class for two time series selected from the `Actor' group. The bi-modal distribution for Glenn Close indicates that the model has recognized the fact that she is both an actress and a singer.}
                 \label{fig:post}
            \end{figure}
            
            \begin{figure}
               \begin{center}
                \includegraphics[width=150mm]{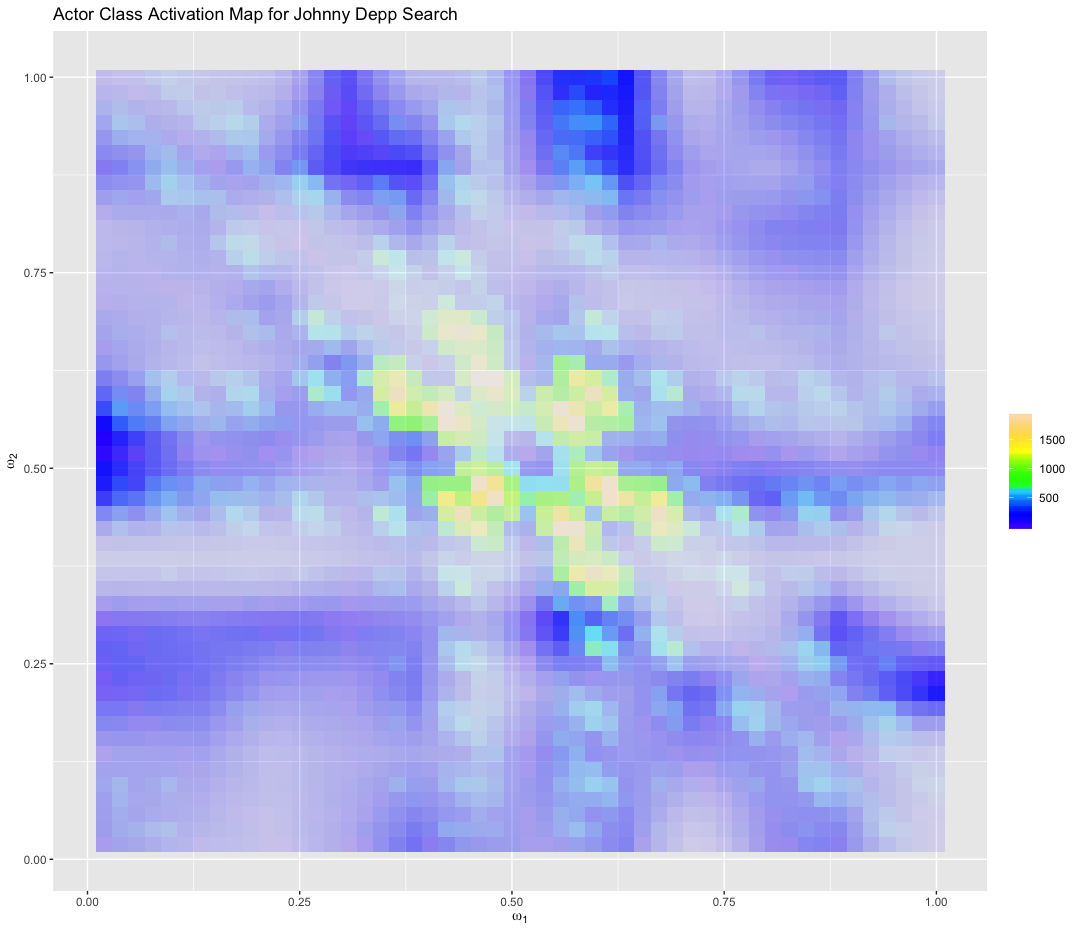}
                \end{center}
                 \caption{\baselineskip=10pt Class activation map overlaid on original bispectral image for Johnny Depp search. The CAM is visualized via transparency, where more transparent areas are less important, and more vivid regions are more important.}
                 \label{fig:cam}
            \end{figure}

    Typically in a classification problem such as this, one would split the data into a training, validation, and test set. The model would be trained on the training set, and any tuning parameters would be chosen based on out of sample prediction using the validation set. Finally, the test set would be used to gauge out of sample performance (see \citet{draper13} for examples of this framework). Because we have a limited sample size, we compare the proposed BCNN to spectral-SSVS using leave-$k$-out cross validation. For each iteration of the cross-validation routine, we randomly sample $k$ time series as a validation set and another $k$ as a test set, thus leaving the remaining time series for training. We fit the model using the training set, and make predictions on both the validation set and test set. Finally, we compare our predicted vs. true responses over all iterations of the cross-validation routine. Our dropout rate was chosen based on the validation results and the test set results were used to estimate out of sample accuracy and area under the ROC curve (AUC). In this example, we use $k=10$ and use 50 iterations of cross-validation, resulting in 500 total predictions each for the validation and test sets.
    
    We use the same BCNN architecture as in our simulation study. The only parameter we tune is the dropout rate, where we tune using grid search over the values $(0.02, 0.04, 0.06, 0.08, 0.1)$. We also use the same spectral-SSVS settings as in the simulation.  For spectral-SSVS, we attain out of sample accuracy of 0.556 and AUC of 0.626. This indicates that the second moment properties are likely not sufficient for classifying these time series. However, our best results for BCNN, using a dropout rate of 0.1, yielded an accuracy of 0.634 and an AUC of 0.649. BCNN is more successful than spectral-SSVS at this difficult classification problem, even considering the limited sample size.  The third moment properties add critical information about this problem, and the BCNN works well by providing dimension reduction of the bispectra, while simultaneously modeling complex nonlinear functions.
    
    In Figure \ref{fig:post}, we show the posterior distributions of the probability that the time series is from an `actor' search term, for two selected `actor' time series, after fitting on the full dataset. Notice that for the `Glenn Close' time series, there is a bi-modal distribution, with one mode above 0.5 and one mode below 0.5. A quick internet search reveals that Glenn Close is well known for both acting and singing. In this case, the uncertainty quantification provided by dropout adds insight that would not be available otherwise.
    
    Along with uncertainty quantification, another benefit of the BCNN is inference. Because the BCNN relies on a convolutional neural network structure, a technique known as class activation mapping (CAM) may be used \citep{selv17}. The general idea behind CAM is to generate a heatmap of which regions in a given image are most significant in determining whether or not the image comes from a specific class. This is accomplished by taking the final convolution layer and weighting each filtered image by the gradient of the class with respect to the filter. We present an example CAM for the actor class in Figure \ref{fig:cam}. There are a few regions of importance in determining the actor class, such as the frequency pairs around 0.5 and 0.1.

	\subsection{Electric Devices Data}

	As another example comparing BCNN to spectral-SSVS, we use the electric devices data from \citet{lines11}, obtained from the UEA and UCR Time Series Classification Repository \citep{bagnall18}. The data consists of electricity consumption time series of length 96, coming from 7 types of household electric devices. The data contains 8,926 time series in the training set, and 7,711 in the test set. We subset the data to only include the first two device types (television and computer), resulting in a training set of size 2,958 and a test size of 2,623. The goals is to classify the time series into the correct device type grouping based on the electricity consumption time series.
	
	We use the same network structure as in the Google Trends example. Because this dataset is much larger, we only fit the model for two epochs, using a batch size of 128. The data includes a predefined test set, so rather than use a cross-validation procedure, we just compare BCNN vs. spectral-SSVS on the test set after fitting on the training set. Spectral-SSVS (keeping all principal components) yields a classification accuracy of 0.63 and an AUC of 0.71. BCNN using a dropout rate of 0.1 improves these results considerably, with an accuracy of 0.92 and an AUC of 0.83. This is a 46\% and 17\% increase respectively.
	
	In this example, we also compare to variable selection of the spectral principal components using Lasso with a binomial response \citep{tib96}. We use the glmnet R package \citep{friedman2009} for this, which selects the optimal shrinkage penalty via cross-validation. This results in a classification accuracy of 0.65 and an AUC of 0.73. This method is very similar in nature to spectral-SSVS, so it is not surprising that the two achieve similar results.  The Bayesian nature of spectral-SSVS gives it some advantage regarding uncertainty quantification, however, we choose to compare to Lasso here, as software is readily available that will extend Lasso to the multiclass (Multinomial) case.
	
\begin{figure}
                \begin{center}
                \includegraphics[width=150mm]{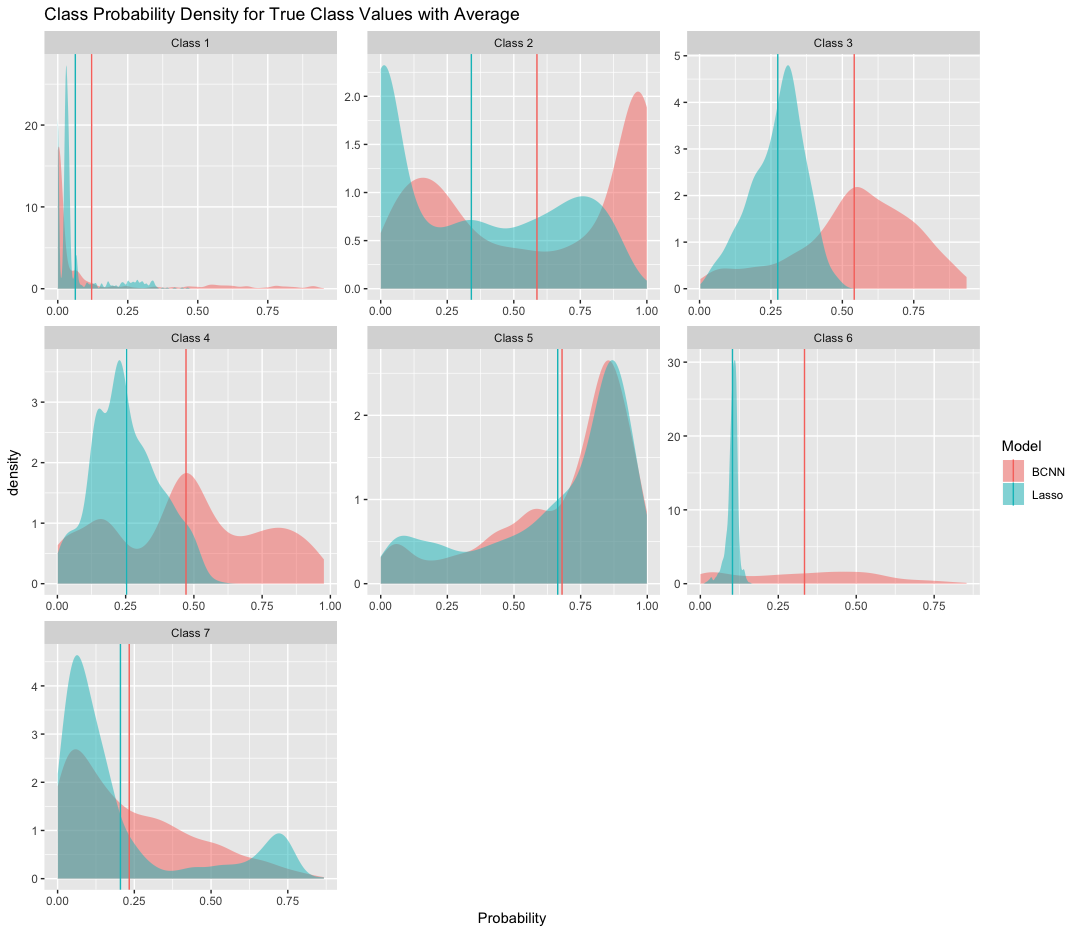}
                 \caption{\baselineskip=10pt Density of class probability under BCNN and spectral-Lasso. Each frame shows the density of all model class probabilities for time series from the given class. A model with more mass shifted towards one indicates superior performance compared to a model that has mass shifted towards zero. The averages are overlaid with vertical lines.}
                 \label{fig:probDens}
 \end{center}               

            \end{figure}

	We further compare BCNN on the full electric device dataset (using all 7 categories) to spectral-Lasso with a multinomial response in order to assess the multiclass classification ability of our methodology. For the multiclass case, we still only include one dense hidden layer, but we increase the number of units in this layer to 32. We chose to use a droput rate of 0.1, but this value could be further tuned. In this scenario, BCNN classifies 58\% of the test cases correctly, whereas spectral-Lasso only classifies 44\% correct. For reference, the naive approach of classifying each time series into the most often observed class achieves a total classification accuracy of 0.24. To further illustrate the advantage of using bispectral properties for time series classification, we present the class probability densities under each model in Figure \ref{fig:probDens}. Each frame in the figure considers only the test set time series from the corresponding class label. Within each frame, we present the density of the corresponding class probability point estimates. An ideal model would have all the mass shifted to one in each frame. Here, we see that both models behave similarly when the true class is either 5 or 7, but BCNN outperforms spectral-Lasso in every other case.

\section{Discussion}\label{sec:disc}

We present a powerful and interpretable model for time series classification that relies on the construction of the sample bispectrum.  The bispectral images are used as inputs into a deep convolutional neural network. CNNs excel at image analysis, because they are translation invariant, reduce dimensionality, and may find many important features within an image.  We find that our BCNN may lead to more accurate predictions than methods that rely only second order spectral properties. For many nonlinear time series, third order properties can be critical, and BCNN successfully handles these cases. Furthermore, the purpose of our applications was to illustrate our methodology, and thus we did consider variations of many of the model choices. Cross-validation over architechure decisions such as the number size of layers could likely yield even higher predictive accuracy.

Inference is another benefit of BCNN. By using parameter dropout as a regularization procedure both during model training and prediction, BCNN allows for theoretically justified uncertainty estimates. In many ways, this is similar to a posterior distribution achieved from Bayesian model averaging, albeit obtained from a complex nonlinear model. Inference can further be made through the use of CAMs, which illustrate the regions of a given image that were important in the determination of a given class prediction. These two advantages help to bridge the gap between machine learning methods which may not always account for uncertainty and statistical methods which do.

Further work may involve adding a time varying component to the bispectrum construction. The BCNN currently makes an implicit assumption that time series are stationary, but by including a time varying component, we could introduce the idea of nonstationarity. The corresponding deep model may include some type of recurrent network structure to account for dependence in time. Additional work may be done by extending the BCNN structure to be used for clustering rather than classification.

\section*{Acknowledgement}

This research was partially supported by the U.S. National Science Foundation (NSF) under NSF SES-1853096 and through the Air Force Research Laboratory (AFRL) Contract No. 19C0067.  This article is released to inform interested parties of research and to encourage discussion.  The views expressed on statistical issues are those of the authors and not those of the NSF or the U.S. Census Bureau.

\bibliographystyle{apacite}
\bibliography{bispecBib}

\end{document}